\documentclass[11pt]{article}

\usepackage[preprint]{acl}

\usepackage{times}
\usepackage{latexsym}

\usepackage[T1]{fontenc}

\usepackage[utf8]{inputenc}

\usepackage{microtype}

\usepackage{inconsolata}

\usepackage{graphicx}

\usepackage{amsfonts}
\usepackage{amsmath}
\usepackage{amssymb}

\usepackage{tabularx}
\newcolumntype{Y}{>{\centering\arraybackslash}X}

\newcommand{\modelname}{CoLT}

\title{CoLT: Reasoning with Chain of Latent Tool Calls}

\author{Fangwei Zhu\thanks{Work done during internship at Bytedance BandAI.}, Zhifang Sui \\
  School of Computer Science, \\
  State Key Laboratory of Multimedia Information Processing, \\ 
  Peking University \\
  \texttt{zhufangwei2022@stu.pku.edu.cn} \\
  \texttt{szf@pku.edu.cn} \\
}

\begin{document}
\maketitle
\begin{abstract}
Chain-of-Thought (CoT) is a critical technique in enhancing the reasoning ability of Large Language Models (LLMs), and latent reasoning methods have been proposed to accelerate the inefficient token-level reasoning chain.
We notice that existing latent reasoning methods generally require model structure augmentation and exhaustive training, limiting their broader applicability.
In this paper, we propose CoLT, a novel framework that implements latent reasoning as ``tool calls''.
Instead of reasoning entirely in the latent space, CoLT generates seed tokens that contain information of a reasoning step.
When a latent tool call is triggered, a smaller external model will take the hidden states of seed tokens as its input, and unpack the seed tokens back to a full reasoning step.
In this way, we can ensure that the main model reasons in the explicit token space, preserving its ability while improving efficiency.
Experimental results on four mathematical datasets demonstrate that CoLT achieves higher accuracy and shorter reasoning length than baseline latent models, and is compatible with reinforcement learning algorithms and different decoder structures.
\end{abstract}

\section{Introduction}

Large Language Models (LLMs) have shown impressive capabilities on various reasoning-centric tasks~\cite{2025gpt5, 2025gemini, 2025claude} by adopting Chain-of-Thought (CoT) prompting techniques~\cite{wei2023chain}.
CoT breaks down complex problems into simpler steps, and has been proven to be crucial to solve problems that require serial multi-step reasoning~\cite{feng2023towards, li2024chain}.
However, the explicit CoT approach needs to generate the reasoning chain token by token, and may face high computational costs during inference, which poses a challenge to real-world LLM applications.

To pursue efficient reasoning, implicit CoT approaches~\cite{hao2024training, shen2025codi, tan2025think} have since been proposed by guiding models to reason in a continuous latent space rather than the token space.
As a representative example, COCONUT~\cite{hao2024training} iteratively replaces reasoning steps with latent tokens, whose output state is directly used as subsequent input embeddings.
More recent approaches like CODI~\cite{shen2025codi}, COLAR~\cite{tan2025think} and SIM-CoT~\cite{wei2025sim} further introduce more signals like trajectory-level alignment, token chunk distributions or step-level supervision to help LLMs better learn from existing reasoning chains.
While effectively reducing the length of CoT, these methods generally require major modification to model structure and intensive training, restricting their practical employment across different tasks.

Recently, \citet{zhu2025chain} suggests that CoT tokens play the role of storing intermediate results and the results can be represented in different forms.
Deepseek-OCR~\cite{wei2025deepseek} also demonstrates that a single vision token may contain sufficient information to be decoded into multiple tokens.
Inspired by these discoveries, we introduce \modelname{} (\textbf{C}hain-\textbf{o}f-\textbf{L}atent-\textbf{T}ools), a framework that implements latent reasoning as parametric tool calls.

Rather than reasoning entirely in the black-box latent space, \modelname{} guides LLMs to generate special ``seed tokens'' that contain condensed information of a reasoning step in their hidden states.
When the main model raises a tool call, \modelname{} will invoke an external decoder to decode the hidden states of the seed tokens back to normal text tokens, and the main model will continue its reasoning with the new context.
Unlike traditional tools that are discrete APIs, we implement the decoders as differentiable neural modules.
When \modelname{} offloads the expansion of reasoning steps to the decoders, the gradients could be used to jointly optimize the main model.

In this way, we model the latent reasoning process as a parametric tool call that can be combined with different decoders, which anchors the main model in the pretrained text space.
By forcing the model to generate condensed hidden states and decode them back to explicit text tokens, we ensure the main model could reason with text token contexts, thus maintaining the interpretability of reasoning chains.
Meanwhile, when equipped with decoders that support sampling, \modelname{} is naturally compatible with reinforcement learning algorithms without the need to manually inject noise, allowing it to better explore the right reasoning path.

We evaluate \modelname{} on four grade-school mathematical reasoning datasets: GSM8k~\cite{cobbe2021training}, GSM8k-hard~\cite{gao2022pal}, SVAMP~\cite{patel2021nlp} and MultiArith~\cite{roy2015solving}.
Experimental results show that \modelname{} outperforms existing methods without the need for training for many epochs and could further benefit from reinforcement learning on more difficult datasets.

We further investigate the performance of different decoder structures.
While decoders with the Transformer structure achieve the best performance, there also exist alternative choices like multi-hot decoder, revealing the potential of expanding the \modelname{} framework to broader realms.

To sum up, our contributions are as follows:
\begin{itemize}
    \item We propose \modelname{}, a novel framework that implements latent reasoning as tool calls to decode text from special seed tokens, which accelerates reasoning while preserving the pretrained ability of reasoning in the text space.
    \item We design different decoders for \modelname{}, demonstrating the influence of decoder parameters and their potential of expanding to other forms.
    \item We evaluate \modelname{} on four benchmarks, proving its outstanding ability over previous methods. Experiments on more difficult benchmarks further demonstrate its compatibility with reinforcement learning.
\end{itemize}

\section{Related Work}
\subsection{Explicit and Implicit CoT Reasoning}
Chain-of-Thought (CoT) reasoning~\cite{wei2023chain} is a core technique in enhancing the reasoning abilities of large language models (LLMs).
By generating intermediate reasoning steps, LLMs have achieved success in various tasks~\cite{guo2025deepseek}, and reinforcement learning techniques~\cite{guo2025deepseek, 2025gpt5, team2025kimi} further pushed the capability boundaries of LLMs in verifiable problems.
To reduce the heavy computational cost of long CoTs, various token-level optimization solutions have been proposed, for example skipping unimportant tokens~\cite{xia2025tokenskip} or adopting compact forms like mathematical expressions or code~\cite{xu2025chain}.

Another line of research is to leave the explicit token space and reason in the implicit vector space.
Early works like iCoT~\cite{deng2024explicit} attempts to gradually internalize each reasoning step, while pause tokens~\cite{goyal2023think} and filler tokens~\cite{pfau2024let} provide models with special tokens for extra reasoning.
Coconut~\cite{hao2024training} proposes the concept of latent reasoning, which reasons with transmission of hidden states rather than explicit tokens.
Coconut obtains supervision signals by replacing explicit reasoning steps with consecutive latent tokens step by step, and CODI~\cite{shen2025codi} performs self-distillation on the hidden states of the last token to ensure consistency.
More recent methods like COLAR~\cite{tan2025think} and SIM-CoT~\cite{wei2025sim} generally follow the paradigm of compressing explicit tokens into consecutive latent tokens, while introducing additional supervision signals like token block embedding distribution or step-level token supervision to improve stability.

\begin{figure*}
    \centering
    \includegraphics[width=0.9\linewidth]{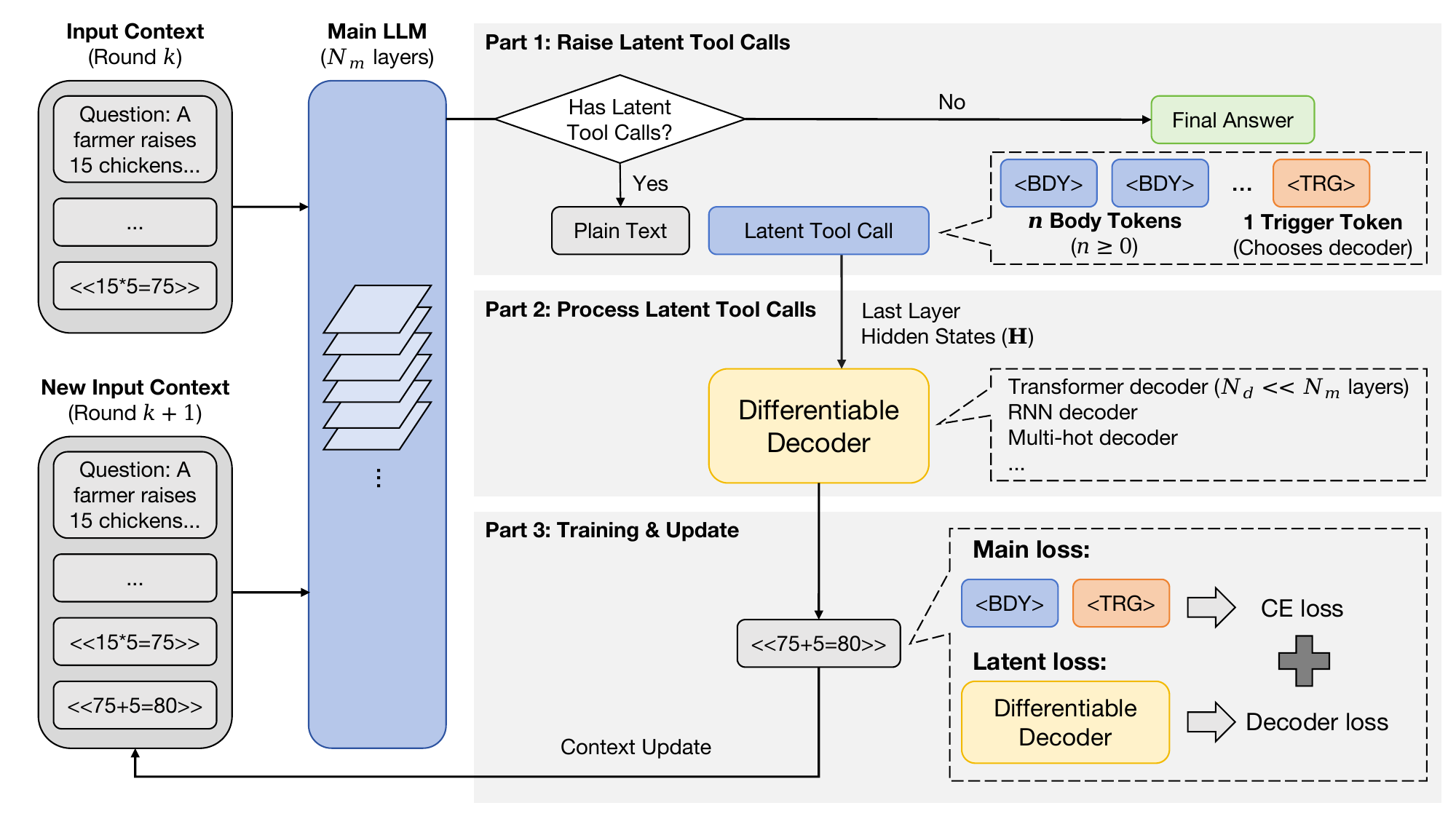}
    \caption{The \modelname{} framework with a standard LLM backbone and differentiable decoders. During inference, the backbone LLM can raise a latent tool call consisting of special body tokens \texttt{<BDY>} and trigger tokens \texttt{<TRG>}. \modelname{} will select an external decoder according to the trigger token to decode the hidden states of the latent tool call back to normal tokens. The LLM will continue its reasoning with the updated context until it reaches the final answer.}
    \label{fig:main_datapath}
\end{figure*}

\subsection{Explaining and Compressing CoT Tokens}
There have been theoretical proofs on the necessity of CoTs in complex problems~\cite{feng2023towards, li2024chain}: the algorithmic complexity that an LLM can solve in a single step has an upper bound, and CoT could raise the upper bound to a higher magnitude.
Recurrent Transformers could be Turing complete~\cite{giannou2023looped}, and Transformers equipped with CoT have similar capabilities~\cite{xu2025cot}.
A hypothesis is that certain tokens in CoT play the role of storing intermediate results, and as long as the critical results are preserved, the CoT could be represented in different forms~\cite{zhu2025chain}.
In this way, the core challenge in latent reasoning is to preserve the necessary information.

Deepseek-OCR~\cite{wei2025deepseek} shows it is possible to compress text tokens in a few visual tokens and then unpack them with marginal information loss, which inspires the analogy that LLM itself may also be able to generate small ``packs'' that contain sufficient information for unpacking into original texts.
We also draw inspiration from speculative decoding~\cite{leviathan2023fast}, which reveals that it is possible to decode tokens for a large LLM with smaller models.

\section{Method}
Figure \ref{fig:main_datapath} shows the overall structure of our proposed framework \modelname{}, which consists of two modules: a main LLM $M$ and a set of decoders $\mathcal{D} = \{D_1, D_2, \ldots, D_n\}$.
Given a question $Q$, the main LLM will go through a multi-step reasoning chain $R$, and finally reach an answer $A$.

\modelname{} reasons in a ``parametric tool call'' style: it uses differentiable neural modules to replace the non-differentiable traditional tools.
During the reasoning chain, the main model could generate seed tokens to \textbf{raise} a latent tool call, effectively compressing relevant information in the hidden states of seed tokens.
A corresponding decoder $D_i$ will then \textbf{process} the tool call, unpacking the hidden states back to explicit text tokens.
The decoded text tokens will be concatenated to the reasoning chain as the new input context, and this tool call loop continues until the final answer is reached and no more tool calls are raised.

By offloading the decoding task, we could allow the main model to maintain its textual reasoning ability, while ensuring that the entire process remains end-to-end differentiable.

\subsection{Raising Latent Tool Calls}
Previous research on latent reasoning~\cite{hao2024training} has shown that multiple text tokens can be compressed into a single latent token.
\modelname{} utilizes seed tokens to store the latent information.
As illustrated in Figure \ref{fig:main_datapath}, there exist two types of seed tokens: body tokens (\texttt{<BDY>}) and trigger tokens (\texttt{<TRG>})\footnote{We use existing special tokens as seed tokens in our experiments. See Appendix \ref{sec:app_special_tokens} for details.}.

A representative reasoning step that raises a latent tool call would be like:
\begin{verbatim}
    [Q] [PR] <BDY> ... <TRG> <eos>
\end{verbatim}
where \texttt{[Q]} represents the question, \texttt{[PR]} represents previous decoded reasoning steps, and \texttt{<eos>} represents the end-of-sequence token.

The number of consecutive body tokens is optional and is usually viewed as a hyperparameter, but each latent tool call should end with only 1 trigger token.
Body tokens simply act as carriers of latent embeddings, while trigger tokens additionally act as an indicator of which decoder to use.
By constraining the number of seed tokens in a latent tool call, we could force the model to compress its reasoning steps.

We initialize \modelname{} with a token-decoder map $f$, and when a latent tool call ends with a trigger token, we will invoke the corresponding decoder to process the tool call.
The sequence of seed tokens $S = (s_1, \ldots, s_{L_s})$ is viewed as the body of the tool call.
After running the main model $M$, the last-layer hidden states $\mathbf{H} = (\mathbf{h}_1, \ldots, \mathbf{h}_{L_s}) \in \mathbb{R}^d$ of the seed tokens are extracted as seed embeddings.
These embeddings are then passed to the corresponding decoder $D = f(s_{L_s})$ to execute the latent tool call.

After the decoder $D$ processes the tool call and returns decoded text tokens \texttt{[R]}, we will concatenate \texttt{[PR]} with \texttt{[R]} as the new context, and continue autoregressive generation.

\subsection{Processing Latent Tool Calls}
\label{sec:decoders}
After a latent tool call is raised, a latent decoder $D$ processes the tool call by decoding the seed embeddings $\mathbf{H}$ back to textual tokens $R = (r_1, \ldots, r_{L_r})$.
This fully differentiable process allows gradients to backpropagate to the main model, thus enabling joint optimization of the decoder $D$ and the main model $M$.

For the main experiments, we choose Transformer decoders that share the same structure with the main model $M$.
Compared to the main model with $N_m$ layers, the Transformer decoder has significantly fewer layers $N_d \ll N_m$, which leads to highly efficient decoding.
We initialize the parameters of $D$ with the first $N_d$ layers of the main model $M$ to better exploit the low-level features in $M$.

The seed embeddings $\mathbf{H}$ are first projected to the input space of $D$ with a linear projector $P_{D}$, and the decoder $D$ will then autoregressively generate text tokens based on the projected embeddings $\mathbf{Z}$.
The process can be formulated as follows:
\begin{equation}
    \mathbf{Z} = P_{D}(\mathbf{H})
\end{equation}
\begin{equation}
    p_{\phi}(R|\mathbf{Z}) = \prod_{k=1}^{L_r} p_{\phi}(r_k|\mathbf{Z}, r_{<k})
\end{equation}
where $R$ represents decoded text tokens and $\phi$ represents the parameters of the decoder $D$.

The framework also supports alternative decoder implementations, for example projecting $\mathbf{H}$ to number embeddings and decode them back to numbers.
We will further discuss these implementations in Section \ref{sec:decoder_structures}.

\subsection{Supervised Training Objectives}
For each tool call, we first compute the loss over generated tokens to ensure that the main model could correctly raise latent tool calls:
\begin{equation}
    \mathcal{L}_{main} = -\frac{1}{L_s}\sum_{k=1}^{L_s} \log p_{\theta}(s_k|Q, R, s_{<k})
\end{equation}
where $\theta$ refers to the parameters of the main model $M$, $Q$ refers to question tokens, and $R$ refers to tokens in previous reasoning steps.

Afterwards, we calculate the loss of the decoder step.
For Transformer decoders, we calculate the standard cross-entropy loss over all decoded text tokens:
\begin{equation}
    \mathcal{L}_{lat} = -\frac{1}{L_r}\sum_{k=1}^{L_r} \log p_{\phi}(s_k|\mathbf{Z}, r_{<k})
\end{equation}
where $\phi$ refers to the parameters of the decoder $D$, and $Z$ refers to projected seed embeddings.

The final supervised training loss is the sum of two losses:
\begin{equation}
    \mathcal{L}_{sup} = \mathcal{L}_{main} + \mathcal{L}_{lat}
\end{equation}

\begin{table*}[t]
\centering
\scalebox{0.95}{
\begin{tabular}{l|cc|cc|cc|cc|cc}
\hline
    & \multicolumn{2}{c|}{GSM8k-Aug} & \multicolumn{2}{c|}{GSM-Hard} & \multicolumn{2}{c|}{SVAMP} & \multicolumn{2}{c|}{MultiArith} & \multicolumn{2}{c}{Average} \\
    & Acc.          & \#~L          & Acc.          & \#~L           & Acc.         & \#~L         & Acc.           & \#~L           & Acc.           & \#~L \\
\hline
CoT     & 49.4    & 25.6   & 11.9    & 34.2    & 59.8   & 12.1  & 93.2     & 13.7   & 53.6     & 21.4 \\
\hline
iCoT   & 19.8    & 0.00        & 3.87    & 0.00          & 36.4   & 0.00        & 38.2     & 0.00          & 24.6     & 0.00 \\
Coconut & 23.1    & 6.00          & 5.49    & 6.00           & 40.7   & 6.00         & 41.1     & 6.00          & 27.6     & 6.00 \\
CODI & 13.3    & 6.00          & 2.97    & 6.00           & 21.7   & 6.00         & 19.2     & 6.00          & 14.3     & 6.00 \\
COLAR (5x) & 26.8    & 5.57   & 5.87    & 6.53    & 48.4   & 2.95  & 86.4     & 3.21    & 41.7     & 4.57 \\
COLAR (2x) & 40.1    & 12.7   & 9.08    & 14.0    & \textbf{54.9}   & 6.11  & 91.3     & 7.35    & 48.8     & 10.0 \\
SIM-CoT & 42.2    & 13.2   & 9.30    & -    & 43.9   & -   & 87.7     & -    & 47.0     & - \\
\hline
\modelname{} (1seed) & 45.3    & 7.73  & 10.3  & 7.74  & 48.7  & 4.54  &  92.8  & 5.30  & 49.3  & 6.33 \\
\modelname{} (2seed) & \textbf{45.5}    & 10.84  & \textbf{10.8}  & 10.61  & 48.3  & 6.10  &  \textbf{93.9}  & 7.27  & \textbf{49.6}  & 8.70 \\
\hline
\end{tabular}
}
\caption{Experiment results of baseline methods and \modelname{}. \textbf{Bold} text indicates the best result on a benchmark. The results of CoT, iCoT, Coconut, COLAR and CODI are taken from the original paper of COLAR~\cite{tan2025think}, while the result of SIM-CoT is taken from the original paper of SIM-CoT~\cite{wei2025sim} where out-of-domain latent lengths are not reported. The multiplier after COLAR indicates how many consecutive tokens are compressed into a single latent token, which approximates the acceleration ratio.}
\label{tab:main}
\end{table*}

\subsection{Reinforcement Learning with Latent Tool Calls}
By performing supervised training on tool calls constructed from existing reasoning chains, we can teach LLMs to reason with latent tool calls.
To reveal the potential of \modelname{} beyond the golden CoTs, we employ reinforcement learning to encourage exploration of other correct reasoning paths.

As long as the chosen latent decoder $D$ supports sampling, we can view the entire reasoning process as a multi-round conversation.
By sampling tokens from both the main LLM $M$ and the decoder $D$, we can gather diverse reasoning paths for the same question $Q$, and then apply the Group Relative Policy Optimization (GRPO) algorithm to reinforce correct reasoning chains.

To be specific, given a specific question $Q$, we will sample a group of $G$ trajectories $\{T_1, T_2, \ldots, T_{G}\}$ from the old policy $\pi_{\theta_{old}}$.
Each trajectory contains multiple conversation rounds $T_i = \{(P_{i1}, C_{i1}, R_{i1}), \ldots, (P_{iN}, C_{iN}, R_{iN})\}$, where each round consists of a prompt $P$, a main LLM output $C$ and a latent decoder output $R$.
In this way, we can optimize the policy $\pi_{\theta}$ with GRPO by the following objective:
\begin{equation}
    \small
    \begin{split}
        \mathcal{L}_{GRPO} &= -\frac{1}{G} \sum_{i=1}^{G} \sum_{j=1}^{N}(\min(\frac{\pi_{\theta}(C_{ij}, R_{ij}|P_{ij})}{\pi_{\theta_{old}}(C_{ij}, R_{ij}|P_{ij})}A_i, \\
        &\text{clip}(\frac{\pi_{\theta}(C_{ij}, R_{ij}|P_{ij})}{\pi_{\theta_{old}}(C_{ij}, R_{ij}|P_{ij})}, 1-\epsilon, 1+ \epsilon)A_i)) \\
        & +\beta \mathbb{D}_{KL}(\pi_\theta || \pi_{ref})
    \end{split}
\end{equation}

where $\epsilon=0.1$ is a hyperparameter, and $A_i$ is the group averaged reward:
\begin{equation}
    A_i = \frac{r_i - \text{mean}(r_1, r_2, \ldots, r_G)}{\text{std}(r_1, r_2, \ldots, r_G)}
\end{equation}
We set $r_i$ to 1 if the final prediction matches the golden answer, to 0.1 if the answer is wrong but the format is right, and to 0 otherwise.
We also use the KL divergence penalty $\mathbb{D}_{KL}$ with $\beta = 0.05$ to help stabilize the training process.

\begin{figure*}
    \centering
    \includegraphics[width=0.85\linewidth]{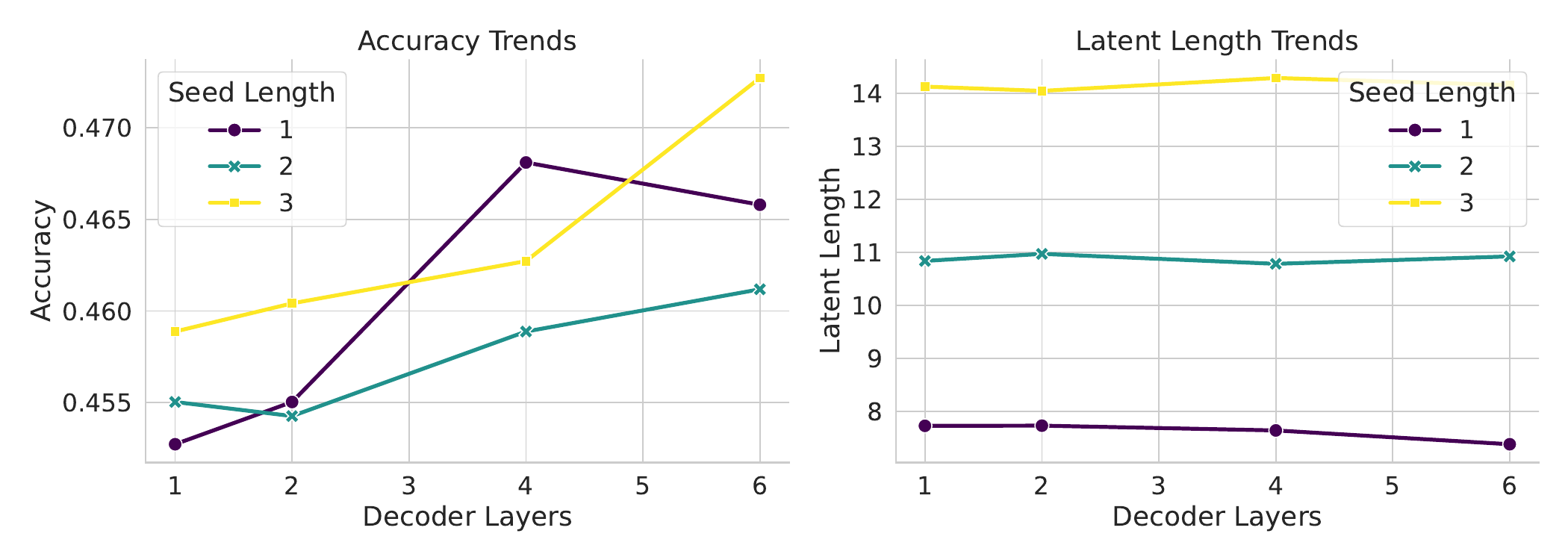}
    \caption{How accuracy and latent length change with the number of seed tokens and decoder layers.}
    \label{fig:decoder_influence}
\end{figure*}

\begin{table*}[t]
\centering
\scalebox{0.95}{
\begin{tabular}{l|cc|cc|cc|cc|cc}
\hline
    & \multicolumn{2}{c|}{GSM8k-Aug} & \multicolumn{2}{c|}{GSM-Hard} & \multicolumn{2}{c|}{SVAMP} & \multicolumn{2}{c|}{MultiArith} & \multicolumn{2}{c}{Average} \\
    & Acc.          & \#~L          & Acc.          & \#~L           & Acc.         & \#~L         & Acc.           & \#~L           & Acc.           & \#~L \\
\hline
\modelname{} (1seed) & 45.3    & 7.73  & 10.3  & 7.74  & 48.7  & 4.54  &  92.8  & 5.30  & 49.3  & 6.33 \\
\hline
RNN & 26.1    & 7.96  & 5.38  & 8.24  & 28.7  & 4.71  &  63.3  & 5.30  & 30.9  & 6.55 \\
Multi-hot* & 17.4    & 39.6  & 4.09  & 42.9  & 26.3  & 20.5  &  65.0  & 22.6  & 28.2  & 32.3 \\
\hline
\end{tabular}
}
\caption{Ablation study results on different decoder structures. *Multi-hot decoders play as a proof-of-concept, they only decode numbers and leave other tokens unchanged, which leads to the much longer \#L.}
\label{tab:ablation_structure}
\end{table*}

\section{Experiments}

\subsection{Experimental Setup}
\paragraph{Training Data.}
Following previous work~\cite{tan2025think}, we use the \textbf{GSM8K-Aug}~\cite{deng2023implicit} dataset to train our model.
GSM8k-Aug is an expanded version of the grade-school level math GSM8k dataset~\cite{cobbe2021training}, consisting of approximately 385k training examples and 1k test samples.
GSM8k-Aug augments the original dataset with GPT-4, replacing natural language reasoning steps with concise mathematical expressions like \texttt{<<25/5=5>>}.
We replace each mathematical expression with a single latent tool call, and view each tool call as a separate training data entry (See Appendix \ref{ssec:app_main_details} for details).

\paragraph{Benchmarks.}
We primarily evaluate models on GSM8k-Aug, which serves as an in-domain benchmark.
We also introduce three additional out-of-domain math reasoning datasets as out-of-domain benchmarks:
(1) \textbf{GSM-Hard}~\cite{gao2022pal}, a modified version of the GSM8k test split by using larger numbers to increase difficulty;
(2) \textbf{SVAMP}~\cite{patel2021nlp} and (3) \textbf{MultiArith}~\cite{roy2015solving} are two simpler grade-school arithmetic word problem datasets.

\paragraph{Evaluation metrics.}
Following previous work~\cite{tan2025think}, we adopt two metrics to evaluate the performance of models: (1) Accuracy (Acc.) that represents the problem-solving ability of models, and (2) Reasoning chain length (\#L) that represents the reasoning efficiency of models.
While the computation cost of decoders is trivial in that they have fewer layers and shorter contexts, considering that we need an extra forward pass to process the new tokens generated by latent decoders, we calculate the length of each latent step in \modelname{} as $L_s + 1$ for fair comparison, where $L_s$ is the number of seed tokens. 

\paragraph{Baseline methods.}
We primarily compare \modelname{} against the following baselines:
(1) \textbf{CoT}~\cite{wei2023chain}, which performs supervised fine-tuning on complete text reasoning chains;
(2) \textbf{iCoT}~\cite{deng2024explicit}, which is a curriculum learning method that gradually removes reasoning steps during training to internalize knowledge, enabling direct prediction during inference;
(3) \textbf{Coconut}~\cite{hao2024training}, which gradually replaces explicit text tokens in the reasoning steps with latent tokens with curriculum learning;
(4) \textbf{CODI}~\cite{shen2025codi}, which self-distills the hidden states across all layers at a selected distillation token to alleviate catastrophic forgetting;
(5) \textbf{COLAR}~\cite{tan2025think}, which compresses consecutive tokens to a vector distribution, enabling sampling and reinforcement learning;
(6) \textbf{SIM-CoT}~\cite{wei2025sim}, which adds step-level supervision with an auxiliary decoder to stabilize optimization. We use the Coconut-based SIM-CoT as baseline.

\paragraph{Implementation details.}
We use Llama-3.2-1B-Instruct~\cite{dubey2024llama} as the backbone of our model.
Unlike previous methods, we train our model from the original backbone model, but we take the reported results of baseline models initialized from SFT-tuned model weights for fair comparison.
We train our model for 2 epochs without curriculum learning, using a fixed learning rate of 1e-5 and a global batch size of 16.
For decoders, we use a minimal 1-layer Transformer decoder with a seed length of 1.
During inference, we set the temperature to 0, i.e. use greedy decoding.
All models are trained on 4 Nvidia GPUs with a memory of 80GB.
For more details, please refer to Appendix \ref{ssec:app_main_details}.

\subsection{Main Results}
Table \ref{tab:main} demonstrates the comparison between \modelname{} and baseline methods on four grade-school math reasoning benchmarks.
It can be clearly observed that \modelname{} outperforms existing latent reasoning methods with only 2 epochs of training.
On the GSM8k-Aug benchmark, \modelname{} achieves a 5\% improvement compared with COLAR (2x), and has a significantly shorter reasoning token length of 7.73 compared with the baseline length of 12.7.

Meanwhile, \modelname{} also shows promising out-of-domain generalization capabilities, achieving superior overall performance, while greatly reducing CoT length.
On the MultiArith benchmark, \modelname{} with 2 seed tokens even surpasses the supervised CoT baseline, proving its great potential.

\begin{table*}[t]
\centering
\begin{tabularx}{0.95\linewidth}{l|YY|YY}
\hline
    & \multicolumn{2}{c|}{DeepSeek-R1-Distill-Qwen-1.5B (MATH)} & \multicolumn{2}{c}{Llama-3.2-1B-Instruct (GSM8k-Aug)} \\
    & Acc.          & \#~L          & Acc.          & \#~L           \\
\hline
CoT* & 23.5    & 209  & 49.4  & 25.6 \\
\hline
\modelname{} & 6.52    & 133.2  & 45.3  & 7.73 \\
\modelname{}+RL & 7.96  & 94.6  & 48.9  & 7.73 \\
\hline
\end{tabularx}
\caption{Reinforcement learning results. We evaluate the effectiveness of reinforcement learning with 2 backbone models: DeepSeek-R1-Distill-Qwen-1.5B on the MATH dataset, and Llama-3.2-1B-Instruct on the GSM8k-Aug dataset. *The CoT results are taken from \citet{tan2025think}.}
\label{tab:reinforce_results}
\end{table*}

\begin{figure*}
    \centering
    \includegraphics[width=0.9\linewidth]{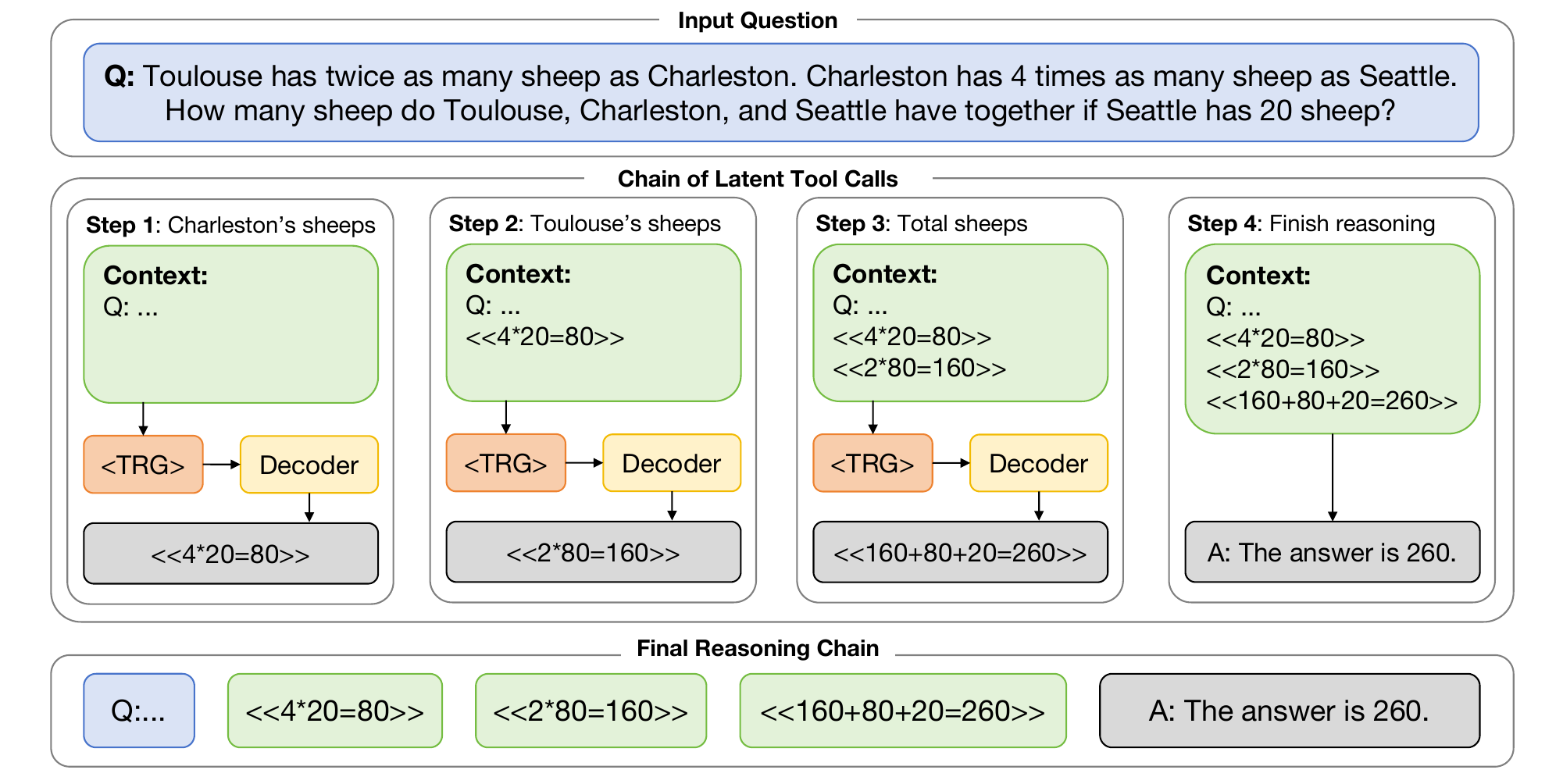}
    \caption{Case study of \modelname{} (1 seed token) on the GSM8K-Aug dataset. For each reasoning step, the main model raises a latent tool call with the \texttt{<TRG>} token, and the decoder decodes the hidden states of seed tokens to equations, which then replaces the seed tokens to form the new context. The input of the main model at each reasoning step is purely textual, and so is the final reasoning chain.}
    \label{fig:case_study}
\end{figure*}

\subsection{Ablation Study}
As stated in Section \ref{sec:decoders}, there exist various implementations for the decoder.
Aside from the minimal 1-layer Transformer decoder with a seed length of 1, we also examine decoders with different number of layers and different seed lengths.
We further implement decoders that go outside the Transformer structure to observe their performance.

\subsubsection{Influence of Layers and Seed Lengths}
Figure \ref{fig:decoder_influence} shows how layer counts and seed lengths influence the performance of latent decoders.
We observe that both the number of decoder layers and seed token length positively contribute to the final accuracy, up to a 2\% accuracy gap between the minimal 1-layer, 1-seed decoder and the 6-layer, 3-seed decoder.
However, considering the computation overhead of larger decoders, the accuracy gain is relatively small, which is the reason why we use a minimal decoder in the main experiments.

For a fixed seed token length, the average reasoning chain length \#L remains stable as the number of decoder layers increases, indicating that the information compressed in a single tool call has little relation with the complexity of decoders.
These results exhibit the potential of achieving better performance by scaling latent decoders.

\subsubsection{Influence of Decoder Structures}
\label{sec:decoder_structures}
We define the role of latent decoders as decoding seed embeddings back to text tokens, which indicates that there exist alternative choices of structures in addition to the Transformer structure.
We construct 2 additional decoders (See Appendix \ref{sec:app_decoder_structure} for implementation details): (1) an RNN decoder that decodes seed embeddings with a simple RNN model, and (2) a multi-hot decoder that maps seed embeddings to multi-hot number vectors.

Table \ref{tab:ablation_structure} shows the performance of alternative decoders.
There exists a clear performance gap between alternative decoders and Transformer decoders, demonstrating the superior performance of the Transformer structure.
Meanwhile, it can be observed that the multi-hot decoder performs poorly, indicating that the recurrent structure is more suited to decode the packed seeds from LLMs.

However, the RNN decoder could still achieve performance similar to baselines like Coconut, and there may exist better decoder structures that remain to be discovered.

\subsection{Reinforcement Learning Results}
To observe the effectiveness of reinforcement learning, we further perform reinforcement learning on GSM8k-Aug with the tuned Llama-3.2-1B-Instruct \modelname{} model.
Additionally, we conduct two-stage SFT-then-RL training on the challenging MATH dataset~\cite{hendrycks2021measuring} by viewing every $k=4$ adjacent tokens as a reasoning step.
We choose DeepSeek-R1-Distill-Qwen-1.5B as the backbone model on MATH to observe the generalization ability of \modelname{} across different backbone models (see Appendix \ref{sec:app_rl_implementation} for implementation details, and Appendix \ref{sec:app_rl_results} for more results).

The experimental results are shown in Table \ref{tab:reinforce_results}.
We can see that performing reinforcement learning on \modelname{} models successfully enhances their performance on both datasets, proving that \modelname{} could learn from exploring diverse reasoning paths.
However, the change in latent length varies, indicating that the base model and the dataset have a major influence on the reinforcement learning process.

It can also be observed that the performance gap between CoT models and \modelname{} models is larger on the MATH dataset, even after reinforcement learning.
In the GSM8k-Aug dataset, there exist clear separations between reasoning steps, while adjacent tokens in MATH do not necessarily have a strong connection between each other, which may be the underlying cause of the performance gap.
Finding clearly defined reasoning steps and converting them into latent tool calls would further benefit the \modelname{} framework.

\subsection{Case Study}
Implicit reasoning models usually lack interpretability as the dense vector space is difficult to align with human-readable text.
\modelname{} alleviates the problem by iteratively replacing latent tool calls with decoded normal text tokens, thus maintaining a readable context.

As demonstrated in Figure \ref{fig:case_study}, each reasoning step is correctly unpacked to normal text tokens, and the final reasoning chain is completely textual.
This proves that \modelname{} is able to perform efficient reasoning while preserving clear reasoning logic.

\section{Discussion}
\paragraph{Exploring better decoder structures.}
In the main experiments section, we prove the effectiveness of Transformer latent decoders.
We further implement 2 additional decoders in Section \ref{sec:decoder_structures}, whose performance falls behind Transformer latent decoders but still exhibit their potential.
In this paper, we focus primarily on compressing multiple text tokens into a single latent tool call.
However, in other tasks such as entity retrieval, decoding the seed tokens with Transformer decoders may not be the best option, and the right decoder structure remains to be explored.

\paragraph{Deciding the input and output boundaries of latent tool calls.}
We train our model on GSM8k-Aug, where the whole reasoning process is clearly divided into equation steps, and thus we can easily define the output of each latent tool call.
For more complex problems, each reasoning step may be longer and exceed the information capacity of seed tokens.
In this way, defining the granularity of each latent tool call, for example sentence-level or token-sequence-level, is a crucial part of applying \modelname{} to more tasks.
Meanwhile, there has been evidence that hidden states in intermediate layers may carry abundant information~\cite{templeton2024scaling}, which reveals the possibility of extracting seed embeddings from non-final layers.

\paragraph{Expanding latent tool calls to other modalities.}
Large language models are trained on text corpus and in this way used to reasoning on text tokens.
For multimodal large language models (MLLMs), they have been trained to perceive visual or audio input data.
BLIP3o~\cite{chen2025blip3} successfully uses hidden states of MLLMs to generate target images.
In this way, it is possible to teach MLLMs to raise latent visual tool calls, where the seed embeddings are passed to an external generator to help MLLMs perform chain-of-thought reasoning in the multimodal space.

\section{Conclusion}
In this paper, we introduce a novel framework \modelname{} that implements latent reasoning in the form of tool calls.
Instead of completely reasoning in the latent space, we train the model to generate seed tokens that contain condensed latent information, and transform the latent information back to normal tokens with differentiable external decoders for future reasoning.

Compared with previous methods, \modelname{} requires less modification to the original model structure, retaining the pretrained ability to reason on text tokens.
Experimental results demonstrate that \modelname{} outperforms existing latent reasoning methods in both accuracy and length of reasoning. 
Meanwhile, when equipped with a decoder that supports sampling, we can model the reasoning process as multi-round conversation and apply existing reinforcement learning algorithms, which improves the performance of DeepSeek-R1-Distill-Qwen-1.5B on the MATH dataset.

By constructing alternative decoders, \modelname{} can be applied to various tasks that could exploit latent information such as entity retrieval and multimodal reasoning, which we will explore in the future.

\section*{Limitations}
In this paper, we focus mainly on math reasoning tasks, and the effectiveness of \modelname{} in other tasks remains to be further explored, such as the granularity of latent tool calls and the optimal decoder structure.
Meanwhile, we primarily conduct experiments on small LLMs such as Llama3.2-1B-Instruct, and scaling \modelname{} to larger models may face additional technical challenges.

\bibliography{custom}

\appendix

\section{Special Token Selection}
\label{sec:app_special_tokens}
For convenience, we choose existing special tokens as seed tokens.
For Llama models, we choose \texttt{<|reserved\_special\_token\_0>} as \texttt{<BDY>} and \texttt{<|reserved\_special\_token\_1|>} as \texttt{<TRG>};
For Qwen models, we choose \texttt{<|fim\_prefix|>} as \texttt{<BDY>} and \texttt{<|fim\_middle|>} as \texttt{<TRG>}.

\section{Implementation Details}
\label{sec:app_implementation}

\subsection{Main Experiment Details}
\label{ssec:app_main_details}
\paragraph{Models.}
Following COLAR, we use Llama-3,2-1B-Instruct and DeepSeek-distill-Qwen-1.5B as backbone models.
For Transformer decoders, we use a model that shares the same structure with the backbone model, but with only $N_d$ layers.
The projector $P_D$ for Transformer decoders is a linear layer.
We initialize parameters of the decoder with parameters of the first $N_d$ layers of the backbone model.

\paragraph{Hyperparameters.}
We use the AdamW optimizer with a weight decay of 0 to train our models.
The learning rate is set to 1e-5 for SFT training stage.
The models are trained with a global batch size of 16, and we use greedy decoding (i.e. use a temperature of 0) during inference.
The training process on GSM8K-Aug goes for 2 epochs.

\paragraph{Training details.}
We train the models on 4 NVIDIA GPUs, with the huggingface transformers and accelerate library.
We do not use any early exit strategies, and use the checkpoint at the end of epoch 2 for evaluation.
The \modelname{} models are trained from the original LLM, rather than from finetuned checkpoints.
For the GSM8k-Aug dataset, we construct training data by viewing each equation (for example \texttt{<<25/5=5>>}) as a latent tool call.

\begin{table*}[t]
\centering
\begin{tabularx}{0.95\linewidth}{l|YY|YY}
\hline
    & \multicolumn{2}{c|}{DeepSeek-R1-Distill-Qwen-1.5B} & \multicolumn{2}{c}{Llama-3.2-1B-Instruct} \\
    & Acc.          & \#~L          & Acc.          & \#~L           \\
\hline
CoT* & 23.5    & 209  & 9.71  & 210 \\
\hline
\modelname{} & 6.52    & 133.2  & 2.12  & 169.0 \\
\modelname{}+RL & 7.96  & 94.6  & 3.26  & 42.2 \\
\hline
\end{tabularx}
\caption{Reinforcement learning results on the MATH dataset. *The CoT results are taken from \citet{tan2025think}.}
\label{tab:app_extra_reinforce_results}
\end{table*}

\begin{table*}[t]
\centering
\scalebox{0.95}{
\begin{tabular}{l|cc|cc|cc|cc|cc}
\hline
    & \multicolumn{2}{c|}{GSM8k-Aug} & \multicolumn{2}{c|}{GSM-Hard} & \multicolumn{2}{c|}{SVAMP} & \multicolumn{2}{c|}{MultiArith} & \multicolumn{2}{c}{Average} \\
    & Acc.          & \#~L          & Acc.          & \#~L           & Acc.         & \#~L         & Acc.           & \#~L           & Acc.           & \#~L \\
\hline
CoT     & 49.4    & 25.6   & 11.9    & 34.2    & 59.8   & 12.1  & 93.2     & 13.7   & 53.6     & 21.4 \\
\hline
\modelname{} (1seed) & 45.3    & 7.73  & 10.3  & 7.74  & 48.7  & 4.54  &  92.8  & 5.30  & 49.3  & 6.33 \\
\modelname{} (1seed+RL) & 48.9    & 7.73  & 10.5  & 10.20  & 49.3  & 5.59  & 97.8  & 5.11  & 51.6  & 7.16 \\
\hline
\end{tabular}
}
\caption{Experiment results on grade-school benchmarks after reinforcement learning.}
\label{tab:app_extra_main_results}
\end{table*}

\subsection{Implementation of Reinforcement Learning}
\label{sec:app_rl_implementation}
We use the MATH dataset to perform reinforcement learning for DeepSeek-R1-Distill-Qwen-1.5B.
Unlike GSM8k-Aug, the MATH dataset does not provide an explicit segmentation of reasoning steps.
To address this problem, we split the reasoning chain into token chunks: every $k$ consecutive tokens is viewed as a reasoning step, which resembles multi-token prediction.
In our experiments, we choose $k=4$ and use a 1-seed, 1-layer Transformer decoder, which leads to an approximately $2\times$ speedup.

We first perform supervised fine-tuning on the training set, and then perform reinforcement learning on the tuned model.
For each input prompt, we will perform $G=8$ rollouts, where each sampled trajectory consists of multiple rounds of tool calls.

We adopt LoRA~\cite{hu2022lora} to finetune the main LLM, and finetune the decoder on all parameters.
We set $r = 64$, $\alpha = 128$ and $\text{dropout} = 0.05$ for LoRA finetuning.

For the MATH dataset, we will assign a reward of 1 if the final answer of a trajectory is correct, a reward of 0.1 if the answer is wrong but contains a \texttt{boxed\{\}} answer formatter, and 0 otherwise;
For the GSM8k-Aug dataset, we will assign a reward of 1 if the final answer of a trajectory is correct, a reward of 0.1 if the answer is wrong but contains a ``This answer is:'' formatter, and 0 otherwise.

The format reward will prevent catastrophic forgetting by giving the model a chance to recover when drifting away from correct answer formats.

The reward will be applied simultaneously to all tool call rounds in the trajectory.

\subsection{Reinforcement Learning Details}
\paragraph{Hyperparameters.}
We set $\epsilon=0.1$ and $\beta = 0.05$ in our experiments.

We use the AdamW optimizer with a weight decay of 0 to train our models.
The learning rate is set to 1e-5 on the SFT training stage, and 1e-6 on the RL training stage.

In the SFT stage, the models are trained with a global batch size of 16;
In the RL stage, we perform 8 rollouts for each input prompt with temperature set to 1 and top-p set to 0.9, and train the model with a mini batch size of 4.
We use greedy decoding (i.e. use a temperature of 0) during inference.
The SFT training process on MATH goes for 2 epochs, and RL training process goes for 1 epoch.

\paragraph{Training details.}
We train the models on 4 NVIDIA GPUs whose memories are 80GB, with the huggingface transformers and accelerate library.
The models are trained with bf16 mixed precision in the RL stage for efficiency.

\section{Additional Reinforcement Learning Results}
\label{sec:app_rl_results}
On the MATH dataset, the result of reinforcement learning on Llama-3.2-1B-Instruct is different from the result of DeepSeek-R1-Distill-Qwen-1.5B.
As shown in Table \ref{tab:app_extra_reinforce_results}, the accuracy of Llama-3.2-1B-Instruct increases like DeepSeek-R1-Distill-Qwen-1.5B.
The reasoning chain length of both models is shorter after reinforcement learning, while the length reduction effect on Llama-3.2-1B-Instruct is more obvious.

We hypothesize that the 1-layer decoder is highly sensitive to disturbances, which could be the reason why performing reinforcement learning on MATH has different effects on different models.
These results also reveal that the base model family is important for reinforcement learning.

Table \ref{tab:app_extra_main_results} shows the results of performing reinforcement learning on Llama-3.2-1B-Instruct.
We can see that both the accuracy on GSM8k-Aug and the accuracy on other out-of-domain benchmarks rises, proving that the abilities model gains from reinforcement learning could generalize to other datasets.

\begin{table*}
    \centering
    \scalebox{0.8}{
    \begin{tabular}{c c c}
    \hline
        Dataset/Model Name & Link & License \\
    \hline
        GSM8k-Aug & https://huggingface.co/datasets/zen-E/GSM8k-Aug & Apache 2.0 \\
        SVAMP & https://huggingface.co/datasets/ChilleD/SVAMP & MIT \\
        MultiArith & https://huggingface.co/datasets/ChilleD/MultiArith & - \\
        GSM-Hard & https://huggingface.co/datasets/reasoning-machines/gsm-hard & MIT \\
        MATH & https://huggingface.co/datasets/EleutherAI/hendrycks\_math & MIT \\
        Llama-3.2-1B-Instruct & https://huggingface.co/meta-llama/Llama-3.2-1B-Instruct & llama3.2 \\
        DeepSeek-R1-Distill-Qwen-1.5B & https://huggingface.co/deepseek-ai/DeepSeek-R1-Distill-Qwen-1.5B & MIT \\
    \hline
    \end{tabular}
    }
    \caption{Links and licenses of dataset and models.}
    \label{tab:app_datasets}
\end{table*}

\begin{figure*}
    \centering
    \includegraphics[width=0.85\linewidth]{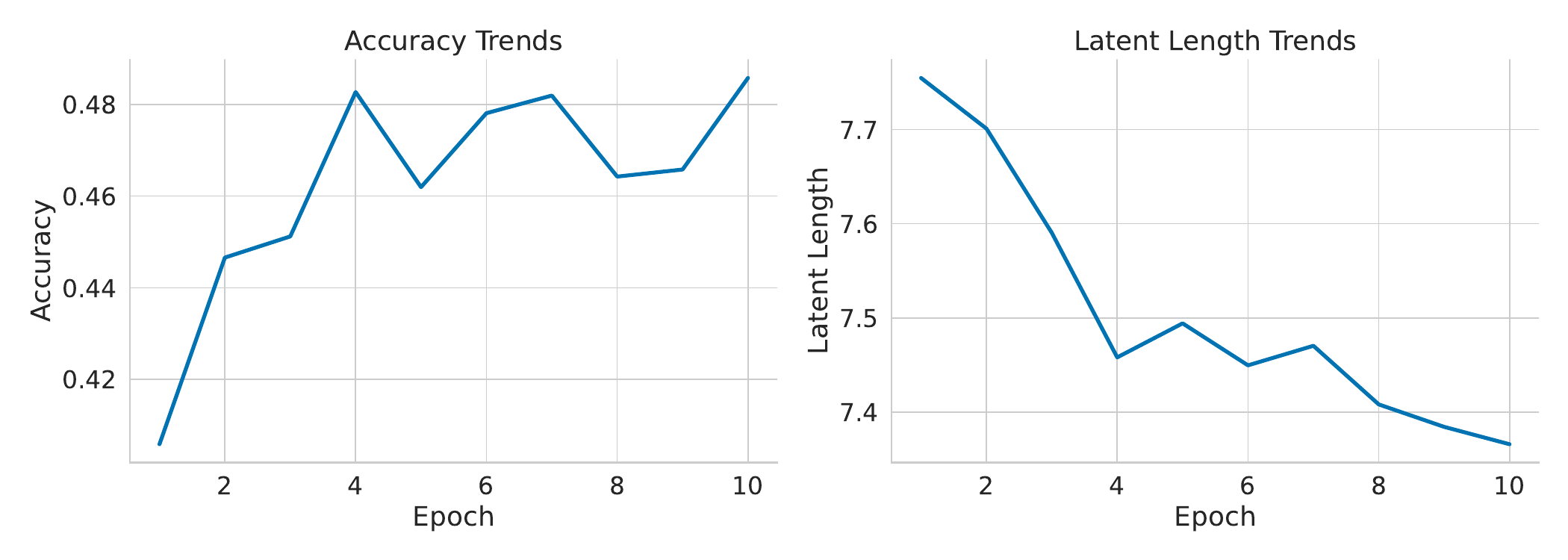}
    \caption{How accuracy and latent length change with the number of training epochs.}
    \label{fig:app_epoch_influence}
\end{figure*}

\section{Alternative Decoder Implementation}
\label{sec:app_decoder_structure}
In Section \ref{sec:decoder_structures}, we construct an RNN decoder and a multi-hot decoder as alternative decoders.

For the RNN decoder, we use a single layer RNN as the model base.
Assume that the decoder takes $n$ seed tokens as input.
At each time step $t$, the input vector $x_t$ equals the projected seed embedding when $t \leq n$, and equals the embedding of the previous token $y_{t-1}$ when $t > n$.
In this way, we can formulate the RNN decoder as follows:
\begin{equation}
  x_t =
  \begin{cases}
    W_{p}\mathbf{H}_t + b_p,  & t \leq n \\
    \text{emb}(y_{t-1}), & t > n
  \end{cases}
\end{equation}

\begin{align}
    h_t &= \text{ReLU}(W_{xh}x_t + W_{hh}h_{t-1} + b_{h}) \\
    y_t &= W_{hy}h_t + b_{y}
\end{align}
where $W_p, b_p, W_{xh}, W_{hh}, b_h, W_{hy}, b_y$ are trainable parameters, $\mathbf{H}$ refers to seed embeddings, and $\text{emb}$ refers to an embedding layer.

For the multi-hot decoder, we use a 2-layer MLP as the model base.
Unlike other decoders, the multi-hot decoder only decodes number values from seed tokens, and the main model generates other textual tokens.
In this way, the reasoning length of the multi-hot decoder is far longer than that of other decoders, and acts only as a proof of concept.

We encode number values as multi-hot vectors $\mathbf{l}$:
a multi-hot vector $\mathbf{l} = (l_1, l_2, \ldots, l_d) \in \mathbb{R}^d$ with $d$ dimensions could represent a number $N$ of at most $n$ digits, where $d = 10n$.
\begin{equation}
    \mathbf{l}_{10k+x} = 
    \begin{cases}
    1, \lfloor \frac{N}{10^{k}} \rfloor \mod 10 = x \\
    0, \lfloor \frac{N}{10^{k}} \rfloor \mod 10 \neq x
    \end{cases}
\end{equation}
Assuming that the value of the $k$-th digit in the little-endian system is $x$, we set $\mathbf{l}_{10k+x} = 1$.

In this way, We can formulate the multi-hot decoder as follows:
\begin{align}
    h = \frac{1}{n}\sum_{i=1}^{n} \mathbf{H}_i \\
    l = Wh + b
\end{align}
where $W, b$ are trainable parameters.
The multi-hot vector $l$ is then converted to numbers, and we tokenize the number as the decoding results.

\section{Effect of Increasing Training Epochs}
Unlike previous methods, we train \modelname{} for only 2 epochs.
Figure \ref{fig:app_epoch_influence} shows how accuracy and \#L change with the number of training epochs.
We can see that the accuracy first rises every training epoch, and stays stable after 4 epochs.
The latent length shares a similar pattern: the length decreases in the first 4 epochs, and stays stable afterwards.

To prevent overfitting, we train the models for 2 epochs in our main experiments.

\section{Usage of AI Assistants}
We use AI assistants to fix grammar errors and polish the writing of this paper.

\section{Licenses of Datasets and Models}
We download all datasets and models from the huggingface hub.
The links and licences are listed in Table \ref{tab:app_datasets}.

\end{document}